\documentclass[pdftex]{article}

\usepackage[final, nonatbib]{neurips_2023}
\usepackage[square,sort,comma,numbers]{natbib}
\usepackage{lipsum}
\usepackage{algorithm}
\usepackage{multirow}
\usepackage{algpseudocode}
\usepackage{amsmath}
\usepackage{amssymb}
\usepackage{natbib}
\usepackage{soul}
\usepackage{algorithm}
\usepackage{adjustbox}

\usepackage[utf8]{inputenc} 
\usepackage[T1]{fontenc}    
\usepackage{hyperref}       
\usepackage{url}            
\usepackage{booktabs}       
\usepackage{amsfonts}       
\usepackage{nicefrac}       
\usepackage{microtype}      
\usepackage{xcolor}         
\usepackage[pdftex]{graphicx} 
\usepackage{microtype}
\usepackage{subfigure}
\usepackage{booktabs} 
\usepackage{color,colortbl}
\usepackage[linewidth=1pt]{mdframed}
\usepackage{framed}
\usepackage{adjustbox}

\usepackage{amsmath}
\usepackage{amssymb}
\usepackage{mathtools}
\usepackage{amsthm}

\usepackage[capitalize,noabbrev]{cleveref}

\theoremstyle{plain}

\theoremstyle{definition}

\theoremstyle{remark}

\usepackage[textsize=tiny]{todonotes}

\title{Mitigating Gradient Overlap in Deep Residual Networks with Gradient Normalization for Improved Non-Convex Optimization}

\author{
  Juyoung Yun$^{1,2}$\thanks{Corresponding Author: \texttt{juyoung.yun@stonybrook.edu}. This work has been accepted to 2024 IEEE International Conference on Big Data Workshop: The Seventh IEEE International Workshop on Benchmarking, Performance Tuning and Optimization for Big Data Applications (BPOD 2024). This paper will appear in proceedings of IEEE International Conference on Big Data 2024.} \\ \\
  $^1$Stony Brook University, Department of Computer Science, USA\\
  $^2$ Stony Brook University, Department of Applied Mathematics and Statistics, USA\\
}

\begin{document}

\maketitle

\let\thefootnote\relax\footnotetext{}

\begin{abstract}
In deep learning, Residual Networks (ResNets) have proven effective in addressing the vanishing gradient problem, allowing for the successful training of very deep networks. However, skip connections in ResNets can lead to gradient overlap, where gradients from both the learned transformation and the skip connection combine, potentially resulting in overestimated gradients. This overestimation can cause inefficiencies in optimization, as some updates may overshoot optimal regions, affecting weight updates. To address this, we examine Z-score Normalization (ZNorm) as a technique to manage gradient overlap. ZNorm adjusts the gradient scale, standardizing gradients across layers and reducing the negative impact of overlapping gradients. Our experiments demonstrate that ZNorm improves training process, especially in non-convex optimization scenarios common in deep learning, where finding optimal solutions is challenging. These findings suggest that ZNorm can affect the gradient flow, enhancing performance in large-scale data processing where accuracy is critical.
\end{abstract}

\section{Introduction}
\label{Introduction}

Deep learning has advanced significantly in recent years, enabling breakthroughs across a wide range of applications. While newer architectures continue to emerge, Residual Networks (ResNets)~\cite{he2016deep} remain widely used in areas like medical imaging and computer vision, valued for their effectiveness~\cite{util1,util2,util3, util4}. ResNets represented a major milestone in deep learning by successfully addressing the vanishing gradient problem through skip connections~\cite{gradientskip}. These connections enable gradients to flow more consistently through deep layers, stabilizing training even in networks with hundreds of layers. This capability has allowed ResNets to outperform traditional models in tasks ranging from image classification to complex data analysis.

Despite their success, We found that the deep residual networks~\cite{he2016deep} have an issue: gradient overlap. This phenomenon results from interactions between gradients from both the learned transformation and the identity mapping (skip connections) during backpropagation. While gradient overlap helps mitigate the vanishing gradient problem by preserving gradient flow, it can also lead to overestimated gradients, resulting in less effective optimization. In particular, during the training, overlapping gradients may cause updates that overshoot optimal regions, potentially slowing convergence. Additionally, in non-convex optimization scenarios, such as those common in deep learning, gradient overlap in ResNets can hinder the model’s ability to find optimal solutions.

\begin{framed}
Can we mitigate the gradient overlap in Residual Networks by applying gradient normalization to control overestimated gradients for the better optimization?
\end{framed}

To address this challenge, our gradient analysis reveals that Z-score Normalization (ZNorm)~\cite{yun2024znorm} can effectively mitigate the effects of gradient overlap. Originally developed as a gradient normalization technique to enhance neural network performance, ZNorm adjusts gradient scales, helping prevent excessive updates by maintaining consistent gradient magnitudes across layers~\cite{yun2024znorm}. This normalization approach not only reduces the risk of overestimated gradients but also accelerating training, aligning gradient scales across layers more effectively.

Our hypothesis is that ZNorm mitigates the gradient overlap phenomenon in ResNets, providing a more balanced and accurate gradient flow that accelerates convergence and enhances training accuracy. This is particularly crucial in fields such as medical imaging, where handling large-scale data accurately is essential, and even minor improvements in model accuracy can have significant implications. By addressing gradient overlap, ZNorm establishes a foundation for understanding why it has proven effective in enhancing performance in CNNs. 

In this paper, we provide the examination of the impact of gradient overlap in ResNets and show how ZNorm offers a solution by stabilizing the gradient flow. Our experimental results demonstrate that applying ZNorm improves optimization stability and leads to more effective deep learning models, particularly in applications requiring high accuracy and robust performance.

\section{Related Works}
Convolutional Neural Networks (ConvNets) have been pivotal in deep learning advancements, especially in computer vision. LeNet-5 by LeCun et al. \cite{lecun1998gradient} laid the groundwork for ConvNets in handwritten digit recognition. The introduction of AlexNet \cite{krizhevsky2012imagenet} by Krizhevsky et al. marked a significant leap in image classification, utilizing deeper architectures and GPU acceleration. Following this, the VGG network \cite{simonyan2015very} by Simonyan and Zisserman explored the impact of network depth, using small convolutional filters and increasing layers to enhance accuracy. However, deeper networks faced challenges like vanishing gradients. To address this, He et al. proposed Residual Networks (ResNets) \cite{he2016deep}, introducing skip connections that allow training of much deeper networks by mitigating these issues. ResNets have become fundamental in various computer vision tasks due to their robust performance and training efficiency. Extensions like Wide ResNets \cite{zagoruyko2016wide}, which trade depth for width, and ResNeXt \cite{xie2017aggregated}, which aggregates transformations, have further improved performance. DenseNets \cite{huang2017densely} enhance feature reuse by connecting each layer to every other layer.

Optimization techniques are fundamental for effectively training deep neural networks. Stochastic Gradient Descent (SGD) and its variants, including momentum-based SGD \cite{qian1999momentum} and adaptive methods like Adam \cite{kingma2014adam}, have been instrumental in minimizing loss functions through iterative parameter updates. Adaptive optimizers such as Adagrad \cite{duchi2011adaptive} and RMSProp \cite{tieleman2012rmsprop} adjust learning rates dynamically based on gradient magnitudes and history, enhancing training stability for deep models. Normalization methods mitigate training instability and smooth optimization landscapes. While Batch Normalization (BN) \cite{ioffe2015batch} normalizes activations to reduce internal covariate shifts, other techniques like Layer Normalization (LN) \cite{ba2016layer} and Group Normalization (GN) \cite{wu2018group} normalize along different dimensions, such as per layer or group of channels. Weight Normalization (WN) \cite{salimans2016weightnorm} and Weight Standardization (WS) \cite{qiao2019weight} focus on consistent scaling of weights, contributing to faster and more stable convergence during training. Gradient adjustment methods stabilize training by controlling gradient updates. Gradient Clipping \cite{clip} caps large gradients, preventing instability. AdamW \cite{adamw} separates weight decay from gradient updates in adaptive optimizers, improving generalization. Stochastic Gradient Sampling focus on improving the generalization effect of Residual Networks~\cite{yun2024stochgradadam}. Gradient Centralization \cite{center} normalizes gradients to zero mean, enhancing convergence. Z-score Normalization (ZNorm) \cite{yun2024znorm} standardizes gradients across layers, mitigating vanishing and exploding gradients for greater stability and performance. In this work, we leverage ZNorm to counteract gradient overlap in ResNets, aiming to further stabilize training by addressing this specific challenge in gradient flow.

\section{Gradient Overlap}
Residual networks~\cite{he2016deep} are well-known for their ability to prevent the vanishing gradient problem using skip connections~\cite{huang2017densely}. These connections improve gradient flow and create a smoother loss landscape~\cite{losslandscape}, leading to more stable training convergence~\cite{gradientskip}. However, our investigation reveals a potential drawback of skip connections: \textit{gradient overlap}. This phenomenon occurs when gradients from the learned transformation and the skip connection combine during backpropagation. This overlap may result in the loss of important gradient details, affecting the network's capacity for precise updates. To analyze this, we present several lemmas that describe gradient behavior in normal networks~\cite{krizhevsky2009learning, krizhevsky2012imagenet} and residual networks~\cite{he2016deep}. We conclude with a theorem demonstrating the impact of gradient overlap in Residual Networks. \\

\noindent\textbf{Lemma 2.1.} Let $y_i = x_{i-1} + f_i(x_{i-1}; W_i)$ represent the output of the $i$-th residual block in a Residual Network~\cite{losslandscape}. The gradient of the loss $L$ with respect to the input $x_0$ is bounded by:
\begin{align}
\left\| \frac{\partial L}{\partial x_0} \right\| \leq \prod_{i=1}^{N} (1 + \alpha_i) \left\| \frac{\partial L}{\partial y_N} \right\|
\end{align}
where $\alpha_i = \left\| \frac{\partial f_i(x_{i-1}; W_i)}{\partial x_{i-1}} \right\|$ is the operator norm of the gradient of the learned function in the $i$-th residual block. \\

\noindent\textit{Proof.} Consider the forward propagation through each residual block~\cite{losslandscape}:
\begin{align}
y_i &= x_{i-1} + f_i(x_{i-1}; W_i), \quad \text{for } i = 1, 2, \dots, N
\end{align}
Here, $x_{i-1}$ is the input to the $i$-th block, and $f_i$ is the residual function of that block. To compute the gradient of the loss function \( L \) with respect to \( x_{i-1} \) during backpropagation, we apply the chain rule:
\begin{align}
\frac{\partial L}{\partial x_{i-1}} &= \frac{\partial L}{\partial y_i} \cdot \frac{\partial y_i}{\partial x_{i-1}}
\end{align}
Since \( y_i = x_{i-1} + f_i(x_{i-1}; W_i) \), the derivative \( \frac{\partial y_i}{\partial x_{i-1}} \) is:
\begin{align}
\frac{\partial y_i}{\partial x_{i-1}} &= I + \frac{\partial f_i(x_{i-1}; W_i)}{\partial x_{i-1}}
\end{align}
where \( I \) is the identity matrix, representing the derivative of \( x_{i-1} \) with respect to itself.

The first term, \( \frac{\partial L}{\partial y_i} \cdot I \), simplifies to \( \frac{\partial L}{\partial y_i} \), and substituting this expression back into the chain rule gives:
\begin{align}
\frac{\partial L}{\partial x_{i-1}} &= \frac{\partial L}{\partial y_i} \cdot \left( I + \frac{\partial f_i(x_{i-1}; W_i)}{\partial x_{i-1}} \right) \\
&= \frac{\partial L}{\partial y_i} \cdot I + \frac{\partial L}{\partial y_i} \cdot \frac{\partial f_i(x_{i-1}; W_i)}{\partial x_{i-1}} \\
&= \frac{\partial L}{\partial y_i} + \frac{\partial L}{\partial y_i} \cdot \frac{\partial f_i(x_{i-1}; W_i)}{\partial x_{i-1}}
\end{align}

Factoring out \( \frac{\partial L}{\partial y_i} \) gives the final expression:
\begin{align}
\frac{\partial L}{\partial x_{i-1}} &= \frac{\partial L}{\partial y_i} \left( I + \frac{\partial f_i(x_{i-1}; W_i)}{\partial x_{i-1}} \right)
\end{align}



\begin{figure*}[hbt!]
\centering
\includegraphics[width=1\columnwidth]{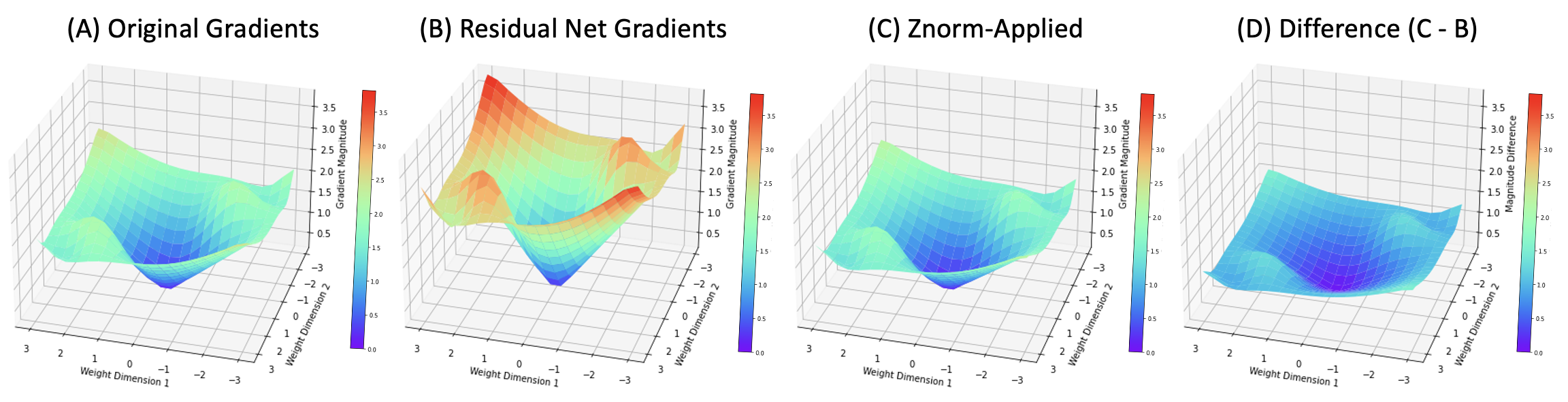}
\caption{Comparison of Gradient Magnitudes in Different Scenarios. This figure visualizes the effects of gradient adjustments in a simulated 3D gradient field, focusing on the impact of residual connections and normalization techniques. Subfigure A displays the original gradients' magnitudes without modifications, representing the initial gradient structure. Subfigure B shows the gradients' magnitudes with residual connections, simulating the effect of gradient overlap often observed in residual networks. Subfigure (C) illustrates the gradients' magnitudes after applying ZNorm~\cite{yun2024znorm}. Subfigure (D) depicts the difference in gradient magnitude between the ZNorm and residual gradients~\cite{he2016deep}. The visualizations emphasize how different normalization techniques and residual structures impact the gradient landscape.}
\label{fig:gradients}
\end{figure*}

Taking the norm on both sides of the inequality and using the sub-multiplicative property of norms, we have:
\begin{align}
\left\| \frac{\partial L}{\partial x_{i-1}} \right\| &= \left\| \frac{\partial L}{\partial y_i} \cdot \left( I + \frac{\partial f_i(x_{i-1}; W_i)}{\partial x_{i-1}} \right) \right\| \\
&\leq \left\| \frac{\partial L}{\partial y_i} \right\| \cdot \left\| I + \frac{\partial f_i(x_{i-1}; W_i)}{\partial x_{i-1}} \right\|
\end{align}
Here, we have applied the sub-multiplicative property of norms, which states that for any matrices \( A \) and \( B \), the inequality \( \left\| AB \right\| \leq \left\| A \right\| \cdot \left\| B \right\| \) holds. This results in:
\begin{align}
\left\| \frac{\partial L}{\partial x_{i-1}} \right\| \leq \left\| \frac{\partial L}{\partial y_i} \right\| \cdot \left\| I + \frac{\partial f_i(x_{i-1}; W_i)}{\partial x_{i-1}} \right\|\label{eq:norm_inequality_extended}
\end{align}

Next, to handle the term \( \left\| I + \frac{\partial f_i(x_{i-1}; W_i)}{\partial x_{i-1}} \right\| \), we use the triangle inequality for operator norms. The triangle inequality states that for any two matrices \( A \) and \( B \), we have:
\begin{align}
\left\| A + B \right\| \leq \left\| A \right\| + \left\| B \right\|
\end{align}
Applying this to our case, since the operator norm of the identity matrix \( I \) is 1 (i.e., \( \left\| I \right\| = 1 \)), this simplifies to::
\begin{align}
\left\| I + \frac{\partial f_i(x_{i-1}; W_i)}{\partial x_{i-1}} \right\| &\leq \left\| \frac{\partial f_i(x_{i-1}; W_i)}{\partial x_{i-1}} \right\|  + \left\| I \right\|\\
 &\leq  \left\| \frac{\partial f_i(x_{i-1}; W_i)}{\partial x_{i-1}} \right\| + 1
\end{align}

We define \( \alpha_i = \left\| \frac{\partial f_i(x_{i-1}; W_i)}{\partial x_{i-1}} \right\| \), which represents the operator norm of the gradient of the learned function in the \( i \)-th residual block. Thus, the inequality becomes:
\begin{align}
\left\| I + \frac{\partial f_i(x_{i-1}; W_i)}{\partial x_{i-1}} \right\| &\leq 1 + \alpha_i
\end{align}

Substituting this result back into equation \eqref{eq:norm_inequality_extended}, we obtain:
\begin{align}
\left\| \frac{\partial L}{\partial x_{i-1}} \right\| &\leq \left\| \frac{\partial L}{\partial y_i} \right\| \cdot (1 + \alpha_i).
\end{align}

We can apply this inequality recursively for $i = N, N-1, \dots, 1$. Starting from $i = N$, we have:
\begin{align}
\left\| \frac{\partial L}{\partial x_0} \right\| \leq \prod_{i=1}^{N} (1 + \alpha_i) \left\| \frac{\partial L}{\partial y_N} \right\|
\end{align}

This completes the proof that the gradient norm with respect to the input $x_0$ is bounded by the product of $(1 + \alpha_i)$ terms across all residual blocks. \\

\begin{figure*}[hbt!]
\centering
\includegraphics[width=1\columnwidth]{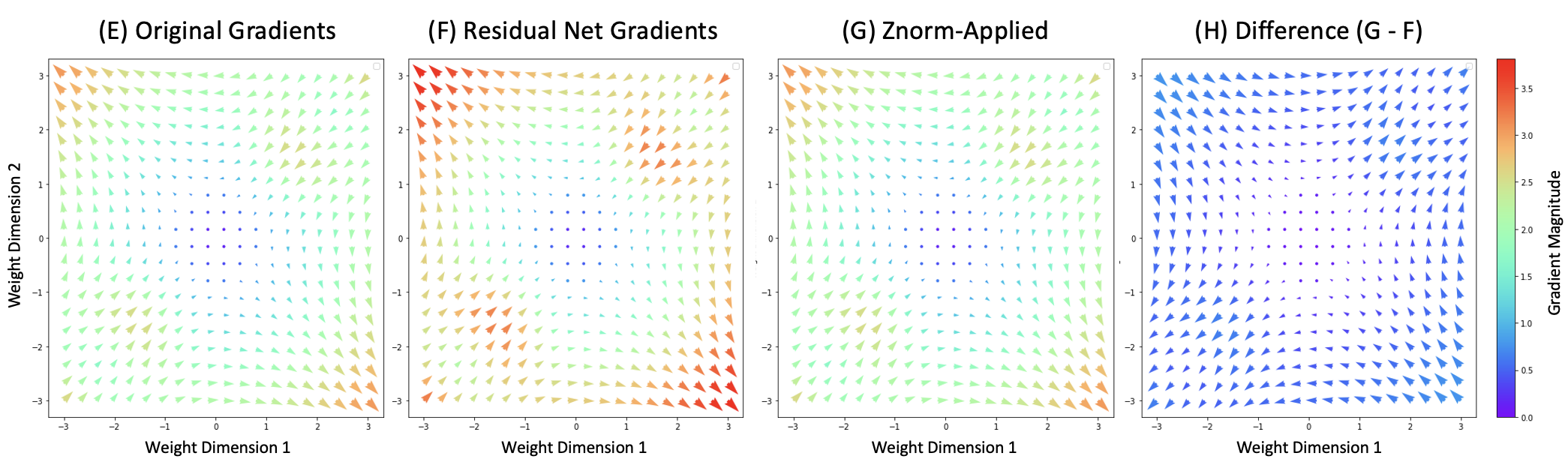}
\caption{Comparison of Gradient Directions Across Various Scenarios. This figure demonstrates how gradient adjustments affect a simulated 3D gradient field, highlighting the role of residual connections and normalization methods. Subfigure A presents the initial structure by displaying the unaltered gradient magnitudes and directions. Subfigure B illustrates the effect of residual connections on gradient magnitudes and directions, replicating the typical gradient overlap seen in residual networks. Subfigure C shows the gradient magnitudes and directions after ZNorm~\cite{yun2024znorm} is applied, and Subfigure D reveals the difference in magnitude and direction between gradients processed with ZNorm~\cite{yun2024znorm} and those from residual connections~\cite{he2016deep}. These visualizations underscore the impact of distinct normalization methods and residual connections on the gradient field.}
\label{fig:gradients2}
\end{figure*}

\noindent\textbf{Lemma 2.2.} For a plain neural network without skip connections, where the output of the $i$-th block is defined as $y_i = f_i(x_{i-1}; W_i)$, the gradient of the loss function $L$ with respect to the input $x_0$ is~\cite{losslandscape}:
\begin{align}
\left\| \frac{\partial L}{\partial x_0} \right\| = \prod_{i=1}^{N} \alpha_i \left\| \frac{\partial L}{\partial y_N} \right\|
\end{align}
where $\alpha_i = \left\| \frac{\partial f_i(x_{i-1}; W_i)}{\partial x_{i-1}} \right\|$ \\

\noindent\textit{Proof.} Consider a plain network without skip connections, where the output of the $i$-th block is:
\begin{align}
y_i = f_i(x_{i-1}; W_i)
\end{align}
For the $i$-th block, the gradient of the loss $L$ with respect to the input $x_{i-1}$ is given by the chain rule:
\begin{align}
\frac{\partial L}{\partial x_{i-1}} = \frac{\partial L}{\partial y_i} \cdot \frac{\partial f_i(x_{i-1}; W_i)}{\partial x_{i-1}}
\end{align}
Taking the norm of both sides:
\begin{align}
\left\| \frac{\partial L}{\partial x_{i-1}} \right\| = \left\| \frac{\partial L}{\partial y_i} \right\| \cdot \left\| \frac{\partial f_i(x_{i-1}; W_i)}{\partial x_{i-1}} \right\|
\end{align}
By setting $\alpha_i = \left\| \frac{\partial f_i(x_{i-1}; W_i)}{\partial x_{i-1}} \right\|$, we have:
\begin{align}
\left\| \frac{\partial L}{\partial x_{i-1}} \right\| = \alpha_i \left\| \frac{\partial L}{\partial y_i} \right\|
\end{align}
Recursively applying this formula for all $N$ blocks, we obtain:
\begin{align}
\left\| \frac{\partial L}{\partial x_0} \right\| = \prod_{i=1}^{N} \alpha_i \left\| \frac{\partial L}{\partial y_N} \right\|
\end{align} \\

\noindent\textbf{Theorem 2.3.} In a Residual Network with \( N \) residual blocks, the presence of skip connections causes \textit{gradient overlap} due to the additive term \( (1 + \alpha_i) \) in the gradient propagation. Specifically, the gradient of the loss function \( L \) with respect to the input \( x_0 \) is bounded by:
\begin{align}
\left\| \frac{\partial L}{\partial x_0} \right\| \leq \prod_{i=1}^{N} (1 + \alpha_i) \left\| \frac{\partial L}{\partial y_N} \right\|
\end{align}
where \( \alpha_i = \left\| \frac{\partial f_i(x_{i-1}; W_i)}{\partial x_{i-1}} \right\| \) is the operator norm of the gradient of the learned transformation in the \( i \)-th residual block. This inequality demonstrates that gradients from both the identity mapping and the learned transformation are combined at each layer, leading to gradient overlap. \\

\noindent\textit{Proof.} From Lemma 2.1, we have established that for each residual block:
\begin{align}
\frac{\partial L}{\partial x_{i-1}} &= \frac{\partial L}{\partial y_i} \left( I + \frac{\partial f_i(x_{i-1}; W_i)}{\partial x_{i-1}} \right)
\end{align}
Taking the norm and using the sub-multiplicative property of norms:
\begin{align}
\left\| \frac{\partial L}{\partial x_{i-1}} \right\| &\leq \left\| \frac{\partial L}{\partial y_i} \right\| \left( \left\| I \right\| + \left\| \frac{\partial f_i}{\partial x_{i-1}} \right\| \right)
\end{align}
Since \( \left\| I \right\| = 1 \), this simplifies to:
\begin{align}
\left\| \frac{\partial L}{\partial x_{i-1}} \right\| &\leq (1 + \alpha_i) \left\| \frac{\partial L}{\partial y_i} \right\|
\end{align}
By recursively applying this inequality from \( i = N \) down to \( i = 1 \), we obtain:
\begin{align}
\left\| \frac{\partial L}{\partial x_0} \right\| &\leq \prod_{i=1}^{N} (1 + \alpha_i) \left\| \frac{\partial L}{\partial y_N} \right\|
\end{align}
This result shows that at each layer, the gradient includes contributions from both the identity mapping (\( 1 \)) and the learned transformation (\( \alpha_i \)), encapsulated in the term \( (1 + \alpha_i) \). This additive term indicates that gradients from these two sources are combined, leading to gradient overlap.

In contrast, for a plain network without skip connections (from Lemma 2.2), the gradient propagation is:
\begin{align}
\left\| \frac{\partial L}{\partial x_0} \right\| &= \prod_{i=1}^{N} \alpha_i \left\| \frac{\partial L}{\partial y_N} \right\|
\end{align}
Here, the gradient at each layer depends solely on the learned transformation's gradient norm \( \alpha_i \).
Comparing the two results, we observe that:
\begin{align}
\prod_{i=1}^{N} (1 + \alpha_i) > \prod_{i=1}^{N} \alpha_i
\end{align}
\(\text{since } 1 + \alpha_i > \alpha_i \text{ for all } \alpha_i > 0\).

The additive \( 1 \) from the identity mapping in residual networks increases the gradient norm compared to plain networks, leading to what we term \textit{gradient overlap}. This overlap can result in an overestimation of the gradient norm, amplifying certain components beyond what is necessary for accurate updates. Although batch normalization~\cite{ba2016layer} helps control the gradient magnitude, it cannot fully mitigate the directional influence of the additive \( 1 \), which can cause a loss of precision in the gradient direction. This loss of gradient directionality can impair optimization efficiency, especially when the learned transformation \( f_i \) is highly sensitive (i.e., when \( \alpha_i \) is large), leading to compounded contributions from both the identity path and the learned transformation. These compounded contributions can influence the optimization process and potentially affect training stability. \\

\noindent\textbf{Gradient Analysis.} Subfigure (A) in Figure~\ref{fig:gradients} and Subfigure (E) in Figure~\ref{fig:gradients2} display the original gradients without residual effects, serving as a baseline. Subfigure (B) in Figure~\ref{fig:gradients} and Subfigure (F) in Figure~\ref{fig:gradients2} demonstrate the influence of skip connections, where the gradients are modified by a residual overlap mechanism. This overlap is simulated by amplifying the existing gradients by a specified fraction, expressed by the equation:
\begin{align}
\frac{\partial Z_{\text{residual}}}{\partial x} &= \frac{\partial Z}{\partial x} + 0.5 \cdot \frac{\partial Z}{\partial x},
\end{align}
where \( \frac{\partial Z}{\partial x} = \left( \frac{\partial Z}{\partial x_1}, \frac{\partial Z}{\partial x_2} \right) \) represents the original gradient components, and the overlap factor (0.5 in this example) controls the degree of gradient amplification. This amplification leads to an intensification of certain gradient directions, as seen in Subfigures (B) and (F), where some gradient vectors are overestimated, potentially resulting in misdirected weight updates. This directional intensification caused by the overlap can disrupt convergence by pushing the model away from optimal gradient descent paths, ultimately impacting training stability and accuracy.

\section{Methodology: ZNorm}
In this section, we discuss Z-score gradient normalization (ZNorm) as an effective technique for managing gradient flow in skip-connected networks. ZNorm~\cite{yun2024znorm} has demonstrated superior performance compared to traditional methods like gradient centralization~\cite{center}, gradient clipping~\cite{clip}, and weight decay~\cite{adamw}. By normalizing gradient magnitudes, ZNorm addresses the accumulation of overlapping gradients in skip connections, leading to more stable and efficient training. \\

\noindent\textbf{Notations.} Consider a deep neural network with \( L \) layers, where each layer \( l \) is associated with weights \( \mathbf{\theta}^{(l)} \). For fully connected layers, the weights are represented as \( \mathbf{\theta}^{(l)}_{fc} \in \mathbb{R}^{D_l \times M_l} \), where \( D_l \) is the number of neurons and \( M_l \) is the input dimension to the layer. In convolutional layers, the weights \( \mathbf{\theta}^{(l)}_{conv} \) are described by a 4-dimensional tensor: \( \mathbf{\theta}^{(l)}_{conv} \in \mathbb{R}^{C_{out}^{(l)} \times C_{in}^{(l)} \times k_1^{(l)} \times k_2^{(l)}} \), where \( C_{in}^{(l)} \) and \( C_{out}^{(l)} \) represent the number of input and output channels, and \( k_1^{(l)} \), \( k_2^{(l)} \) are the kernel sizes. Let \( \nabla \mathcal{L}(\mathbf{\theta}^{(l)}) \) denote the gradient of the loss function \( \mathcal{L} \) with respect to the weights \( \mathbf{\theta}^{(l)} \) of layer \( l \). The overall gradient tensor is represented by \( \nabla \mathcal{L}(\mathbf{\theta}) \). \\

\noindent\textbf{Z-Score Gradient Normalization (ZNorm).} Z-score normalization is applied independently to each layer's gradient \( \nabla \mathcal{L}(\mathbf{\theta}^{(l)}) \), defined as:
\begin{equation}
\Phi(\nabla \mathcal{L}(\mathbf{\theta_{(t)}^{(l)}})) = \frac{\nabla \mathcal{L}(\mathbf{\theta_{(t)}^{(l)}}) - \mu_{\nabla \mathcal{L}(\mathbf{\theta_{(t)}^{(l)}})}}{\sigma_{\nabla \mathcal{L}(\mathbf{\theta_{(t)}^{(l)}})} + \epsilon}
\end{equation}
where \( \mu_{\nabla \mathcal{L}(\mathbf{\theta^{(l)}})} \) is the mean of the gradients in layer \( l \), and \( \sigma_{\nabla \mathcal{L}(\mathbf{\theta^{(l)}})} \) is the standard deviation of these gradients. A small constant \( \epsilon \) (commonly set to \( 1 \times 10^{-10} \)) is added to avoid division by zero.  \\



\noindent\textbf{Effect on Overlapped Gradients.}
In Figure~\ref{fig:gradients} and Figure~\ref{fig:gradients2}, we examine the effect of skip connections and normalization techniques on gradient flow within deep networks. Subfigure (A,E) illustrates the baseline gradients without skip connections, providing a reference for comparison. In contrast, Subfigure (B,F) shows the gradients when skip connections are applied, highlighting gradient overlap, where gradients from both the learned transformation and the skip connection merge. This overlap can lead to overestimated gradients, potentially affecting the gradient descent process by causing larger-than-necessary updates.

To address this, we apply ZNorm to the skip-connected gradients, as shown in Subfigure (C,G). ZNorm effectively reduces the impact of overlapping gradients by re-centering and scaling them, mitigating the overestimation present in Subfigure (B,F). The difference between the ZNorm-applied and skip-connected gradients is visualized in Subfigure (D,H), showing how ZNorm counters the overestimated components and helps maintain gradient direction consistency. As a result, applying ZNorm allows the model to better manage overlapped gradients, leading to more optimal gradient updates and improved training process. This demonstrates that ZNorm can effectively mitigate the gradient overlap problem, enhancing the model's ability to reach optimal solutions.

\section{Experiments}
In this section, we conducted a series of experiments to train Residual Networks using ZNorm method, comparing its performance to other approaches. These experiments aim to test our hypothesis that ZNorm enhances performance in Residual Networks by mitigating gradient overlap, ultimately leading to improved accuracy with optimal directions. \\

\begin{figure*}[hbt!]
\centering
\includegraphics[width=1\columnwidth]{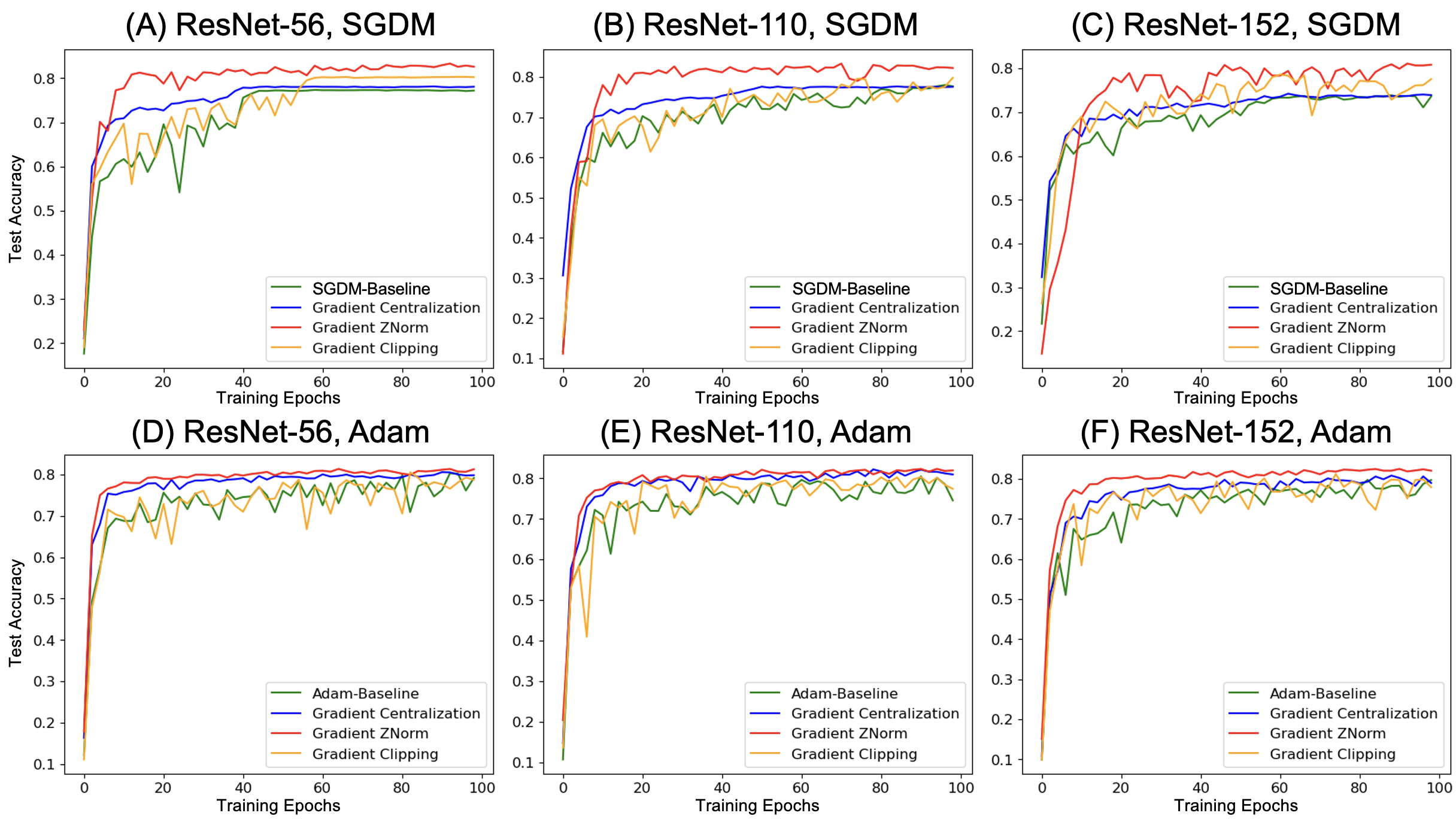}
\caption{Test accuracy comparison on CIFAR-10 Dataset~\cite{krizhevsky2009learning} for Deep Residual Networks~\cite{he2016deep} and gradient normalization techniques~\cite{center, yun2024znorm, clip}.}
\label{fig:plot}
\end{figure*}

\noindent\textbf{Experimental Settings.} All experiments were performed using the Adam~\cite{kingma2014adam} and SGDM~\cite{qian1999momentum} optimizer which has momentum 0.9, with the baseline referring to the standard Adam and SGDM without gradient adjustment. The experiments involved Gradient Clipping~\cite{clip} with 0.1, Gradient Centralization~\cite{center}, and ZNorm~\cite{yun2024znorm}. We train none-augmented CIFAR-10 for 100 epochs with 256 of batch size. \\

\setlength{\tabcolsep}{8pt} 
\begin{table*}[htbp]
\centering
\footnotesize
\caption{Top-1 Performance Comparison on CIFAR-10 Dataset~\cite{krizhevsky2009learning} for Deep Residual Networks~\cite{he2016deep} and gradient normalization techniques~\cite{center, yun2024znorm, clip}. The table shows the test accuracy and train loss across several models and normalization methods. Bold values represent the highest test accuracy achieved for each model.}
\vspace{0.5cm}
\begin{adjustbox}{width=1\textwidth}
\begin{tabular}{@{}l|l|cccc@{}}
\toprule
\toprule
 && \multicolumn{2}{c}{SGDM~\cite{tieleman2012rmsprop}} & \multicolumn{2}{c}{Adam~\cite{kingma2014adam}} \\
\cmidrule(lr){3-4}
\cmidrule(lr){5-6}
Models & Methods & Top-1 Test Accuracy. & Training Loss. & Top-1 Test Accuracy. & Training Loss.  \\
\midrule
\multirow{4}{*}{ResNet-56} & Baseline & 0.7742 &5.2e-05  & 0.8037& 2.0e-03\\
 & Gradient Centralization~\cite{center} & 0.7814 & 6.2e-05 & 0.8129& 1.4e-02\\
 & Gradient Clipping~\cite{clip} & 0.8033 & 4.2e-05 & 0.8054& 1.2e-02\\
 & Z-Score Normalization~\cite{yun2024znorm} & \textbf{0.8331} & 3.6e-02& \textbf{0.8136} & 2.7e-02\\
 \midrule
\multirow{4}{*}{ResNet-110} & Baseline & 0.7802 & 2.0e-04 &0.8131 &1.7e-02\\
 & Gradient Centralization~\cite{center}& 0.7774&7.0e-05 &0.8274 & 1.2e-02\\
 & Gradient Clipping~\cite{clip}& 0.7983 &1.5e-03 &  0.8149& 2.6e-02\\
 & Z-Score Normalization~\cite{yun2024znorm}& \textbf{0.8342}&4.2e-02 & \textbf{0.8280}& 2.1e-02\\
 \midrule
\multirow{4}{*}{ResNet-152} & Baseline& 0.7386 & 4.0e-04 & 0.8004 &1.0e-02 \\
 & Gradient Centralization~\cite{center}& 0.7422 &5.0e-04  & 0.8077 & 2.6e-02\\
 & Gradient Clipping~\cite{clip}& 0.7860 &1.3e-04 &0.8116 & 1.6e-02\\
 & Z-Score Normalization~\cite{yun2024znorm}& \textbf{0.8111} &3.6e-03 &\textbf{0.8267}  & 2.5e-02\\
\bottomrule
\bottomrule
\end{tabular}
\label{tab:accc}
\end{adjustbox}
\end{table*}

\noindent\textbf{Experimental Results.} Figure~\ref{fig:plot} shows the test accuracy progression across 100 epochs for each model and technique. ZNorm consistently led to faster convergence and higher test accuracy compared to baseline and other normalization techniques across both optimizers. Specifically, ZNorm demonstrated its effectiveness in mitigating gradient overlap by maintaining stable gradient flows and minimizing the risk of overestimated gradients, which often disrupt the training in deep residual networks. Table~\ref{tab:accc} summarizes the top-1 test accuracy and training loss achieved by each model and normalization technique. Notably, ZNorm improved both test accuracy validating its capability to mitigate the gradient overlap problem inherent in residual networks. These results confirm that ZNorm not only enhances model performance but also provides a robust solution for managing gradient flow, particularly beneficial in non-convex optimization scenarios typical of deep neural networks.

\section{Conclusion}
Z-score Gradient Normalization (ZNorm) effectively addresses the gradient overlap issue in Residual Networks (ResNets) by standardizing gradient magnitudes across layers, thus minimizing the risk of overestimated gradients from skip connections. This results in more accelerated and optimal training, particularly in deep networks facing non-convex optimization challenges. Experimental results on ResNet architectures showed that ZNorm improves convergence rates and achieves higher test accuracy compared to baseline methods and other gradient normalization techniques. These findings confirm ZNorm’s utility in optimizing ResNet-based models across various domains.

\bibliographystyle{plain}
\bibliography{main}

\end{document}